# A Bayesian Multiresolution Independence Test for Continuous Variables

**Dimitris Margaritis**
Computer Science Dept.
Carnegie Mellon University
Pittsburgh, PA 15213
Dimitris.Margaritis@cs.cmu.edu

**Sebastian Thrun**
Computer Science Dept.
Carnegie Mellon University
Pittsburgh, PA 15213
Sebastian.Thrun@cs.cmu.edu

## Abstract

In this paper we present a method of computing the posterior probability of conditional independence of two or more continuous variables from data, examined at several resolutions. Our approach is motivated by the observation that the appearance of continuous data varies widely at various resolutions, producing very different independence estimates between the variables involved. Therefore, it is difficult to ascertain independence without examining discretized data at several carefully selected resolutions. In our paper, we accomplish this using the exact computation of the posterior probability of independence, calculated analytically given a resolution. At each examined resolution and boundary placement, we assume a multinomial distribution with Dirichlet priors for the discretized table parameters, and compute the posterior using Bayesian integration. Across resolutions, we use a search procedure to approximate the Bayesian integral of probability over an exponential number of possible boundary placements. Our method generalizes to an arbitrary number variables in a straightforward manner. The test is suitable for Bayesian network learning algorithms that use independence tests to infer the network structure, in domains that contain any mix of continuous, ordinal discrete, and categorical variables.

## 1 Introduction and Motivation

Knowledge about the independencies that exist in a domain is a very useful piece of information that a research scientist can—and must—elicit early on during her investigation. Conditional independence statements concerning observed quantities in the domain can greatly aid in the task of understanding the interactions among the domain variables. Imagine for example a medical researcher, attempting to causally model a set of disease and symptom indicators in an effort to find ways of treating or preventing them from occurring in the future. Such conditional independence statements can lend her significant insights to the cause-effect relationships that may be present. Similarly, a solid state physicist attempting to model the interaction among tens of variables representing proportions of substances involved in a new manufacturing process. Knowledge of conditional independencies can help her focus her attention to only the relevant ones during the different stages of the manufacturing process.

More than simply aiding in focusing attention, there exist a number of algorithms today that can induce the causal structure of a domain given a set of conditional independence statements, under assumptions [10, 8]. One principled method for representing such causal structures are Bayesian Networks (BNs) [8], one of the prevalent tools for representing and reasoning about uncertainty. A BN consists of two parts: (1) a directed graphical description of the relationships among the variables in the domain, and (2) a quantitative description of the nature of interactions between each variable and its parents in the graph. In this paper we focus on an integral component of the first part, namely developing a probabilistic conditional independence test for use in causal structure discovery. There exist several approaches for causal discovery in domains with categorical variables [10, 8]. However there are few for continuous or hybrid domains [2, 6]. Most of the latter approaches belong to the score-maximization family of algorithms for structure induction, which does not give any guarantees for producing a correct causal description of the domain. In contrast, the constraint-based family [10, 5] employs conditional independence tests, and has guarantees of inducing the correct causal structure (under assump-tions, most notably the one assuming that all variables are visible). However, to date there is no general approach to constraint-based structure induction for continuous or hybrid domains, mainly due to the difficulty of the general case. Our approach attempts to remedy that, by developing a statistical independence test for continuous variables that places no constraints on the probability distribution of the continuous variables of the domain. In this paper we propose an algorithm which works by eliciting information from continuous data at several resolutions and combining it into a single probabilistic measure of conditional independence. For this purpose, the examination of many resolutions is necessary. An example that helps explain why that is the case is shown in figure 1. Two scatterplots are shown, together with their corresponding discretizations at $3 \times 4$ resolution. The histograms for the two data sets appear similar even though the independence of the axes variables is not. The data on the left show a strong dependence between $X$ and $Y$ while the data on the right plot are independent by construction. In this case, finer resolution histograms have to be examined in order to determine that dependence. This observation is formalized in section 2 below.

At the top level, our approach is formulated as a Bayesian model selection problem, assigning the class of independent distributions a prior probability. This formulation sidesteps the fact that the class of independent distributions lies on a zero-support submanifold in the domain of distribution functions. This is one of the key advantages of the Bayesian approach that our method relies on.

In the next section we present out approach in detail. Our discussion assumes that all variables are continuous. Application to domains that contain ordinal discrete and categorical variables as well as continuous is straightforward and is only briefly described in section 5.

## 2 Method description

As we mentioned above, our method tests for independence at many resolutions. It does so by discretizing the multidimensional space of the variables involved in the test at several resolutions. In this section we present our approach in two stages. In order to explain our method more effectively, we first describe the simple case for testing conditional independence at a single, fixed resolution.

### 2.1 Single resolution, fixed grid test

In this section we are given a $I \times J$ table of counts that represent the discretized scatterplot of variables

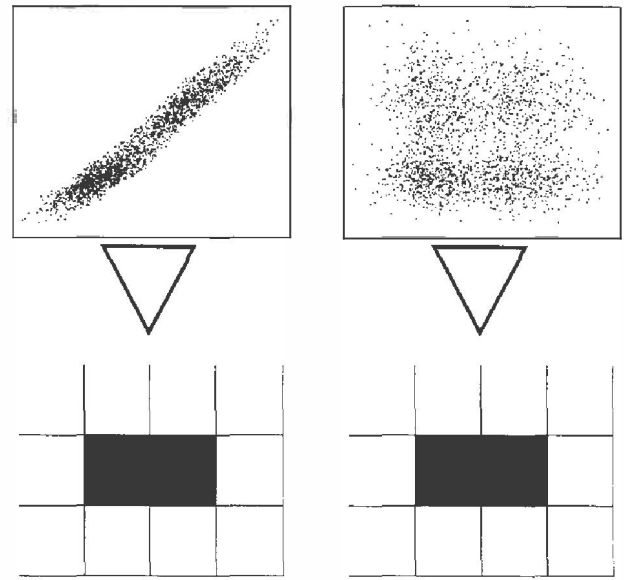

Figure 1: Two very different data sets have very similar $3 \times 4$ histograms. **Left:** Data strongly dependent. **Right:** Data independent by construction.

$X$ and $Y$, and we develop an *exact* test of independence. Such a test has the potential of being used in situations where data are sparse. It can also be used at fine resolutions, a task that will be necessary in the multi-resolution test, described later in section 2.2. We first present the case of an unconditional test of independence ($X \perp Y$) and propose an extension of the method to a conditional test ($X \perp Y \mid \mathbf{C}$) for a given resolution $I \times J \times C_1 \times C_2 \times \cdots \times C_{|\mathbf{C}|}$ near the end of this section.

We assume that the counts of the table, $c_1, \ldots, c_K, K \equiv IJ$, follow a multinomial distribution. The choice of a multinomial is in a sense the most "unbiased" one because it does not make any implicit assumptions about any interaction between adjacent cells (which is sometimes called "smoothing"). We denote the resolution as $R \equiv I \times J$, the set of grid boundaries along the axes as $\mathbf{B}_R$, and the probability of each cell as $p_k, k = 1, 2, \ldots, K$ (for brevity, in the following we denote such a set of numbers as $p_{1 \ldots K}$). The probability of the data set $\mathbf{D}$ (the data likelihood) is the likelihood of the cell counts, namely

$$\begin{aligned} Pr(\mathbf{D} \mid p_{1 \ldots K}, B_R, R) &= Pr(c_{1 \ldots K} \mid p_{1 \ldots K}, B_R, R) \\ &= N! \prod_{k=1}^{K} \frac{p_k^{c_k}}{c_k!} \end{aligned} \quad (1)$$

where $N = |\mathbf{D}|$ is the size of our data set. (For brevity, in the remainder of this section, we omit $B_R$ and $R$ from all probability terms that appear, but assume that they implicitly condition all of them.) Since the parameters $p_k$ are unknown, we adopt a Bayesian approach: we use a prior distribution $Pr(p_{1...K})$ over them. Given such a distribution the data likelihood is the average over all possible parameter values, weighted by their probability:

$$Pr(\mathbf{D}) = \int Pr(\mathbf{D} \mid p_{1...K}) Pr(p_{1...K}) dp_{1...K}. \quad (2)$$

The most commonly used prior distribution for multinomial parameters is the Dirichlet:

$$Pr(p_{1...K}) = \Gamma(\gamma) \prod_{k=1}^{K} \frac{p_k^{\gamma_k - 1}}{\Gamma(\gamma_k)},$$

where $\gamma = \sum_{k=1}^{K} \gamma_k$ and $\Gamma(x)$ is the gamma function.[1] The positive numbers $\gamma_{1...K}$ of this distribution are its hyperparameters.[2] When $\gamma_k = 1$ for all $k$, the Dirichlet distribution is uniform.

The choice of a Dirichlet prior has certain computational advantages. Since it is conjugate prior to the multinomial, the posterior distribution of the parameters is also Dirichlet (albeit with different parameters). This enables us to compute the integral (2) analytically:

$$Pr(\mathbf{D}) = \frac{\Gamma(\gamma)}{\Gamma(\gamma + N)} \prod_{k=1}^{K} \frac{\Gamma(\gamma_k + c_k)}{\Gamma(\gamma_k)}. \quad (3)$$

To test independence, we assume that our data have been produced by one of two classes of models, one representing independence and one that does not. The former one contains $I + J$ parameters for the marginal probabilities of the rows and columns of the $I \times J$ table. We call this model class $M_\mathcal{I}$, and denote its prior probability as $Pr(M_\mathcal{I}) \equiv \wp$. The fully dependent model class, denoted $M_{\neg\mathcal{I}}$, has prior probability $Pr(M_{\neg\mathcal{I}}) = 1 - \wp$. The posterior probability of independence, $Pr(M_\mathcal{I} \mid \mathbf{D})$, is

$$Pr(M_\mathcal{I} \mid \mathbf{D}) = \frac{Pr(\mathbf{D} \mid M_\mathcal{I}) Pr(M_\mathcal{I})}{Pr(\mathbf{D})}$$

by Bayes' theorem. Since

$$Pr(\mathbf{D}) = Pr(\mathbf{D} \mid M_\mathcal{I}) Pr(M_\mathcal{I}) + Pr(\mathbf{D} \mid M_{\neg\mathcal{I}}) Pr(M_{\neg\mathcal{I}})$$

we get

$$Pr(M_\mathcal{I} \mid \mathbf{D}) = 1 \bigg/ \left[ 1 + \frac{(1-\wp)}{\wp} \frac{Pr(\mathbf{D} \mid M_{\neg\mathcal{I}})}{Pr(\mathbf{D} \mid M_\mathcal{I})} \right] \quad (4)$$

---

[1] The gamma function is defined as $\Gamma(x) = \int_0^{+\infty} e^{-t} t^{x-1} dt$. For the case where $x$ is a non-negative integer, $\Gamma(x + 1) = x!$.

[2] The hyperparameters can be thought of as "virtual samples."

At resolution $R$, the term $Pr(\mathbf{D} \mid M_{\neg\mathcal{I}})$ of the fully dependent model that contains $IJ$ parameters is given by (3), i.e.

$$\begin{aligned} Pr(\mathbf{D} \mid M_{\neg\mathcal{I}}) &= \frac{\Gamma(\gamma)}{\Gamma(\gamma + N)} \prod_{k=1}^{K} \frac{\Gamma(\gamma_k + c_k)}{\Gamma(\gamma_k)} \quad (5) \\ &\equiv \Upsilon(\mathbf{c}_K, \gamma_K). \end{aligned}$$

For the independent model we assume two multinomial distributions, one each along the $X$ and $Y$ axes, that contain $J$ and $I$ parameters, respectively. These correspond to the marginal probabilities along the axes. We denote the marginal count at column $j$ as $c_{+j}$ and at row $i$ as $c_{i+}$. The marginal probabilities (which are unknown) are denoted as $q_{1...I}$ with prior a Dirichlet with hyperparameters $\alpha_{1...I}$ and $r_{1...J}$ with hyperparameters $\beta_{1...J}$. The probability of cell $(i, j)$ under $M_\mathcal{I}$ is $q_i r_j$. The data likelihood is computed in a manner analogous to Eq. (3):

$$\begin{aligned} Pr(\mathbf{D} \mid M_\mathcal{I}) &= \left( \frac{\Gamma(\alpha)}{\Gamma(\alpha + N)} \prod_{i=1}^{I} \frac{\Gamma(\alpha_i + c_{i+})}{\Gamma(\alpha_i)} \right) \\ &\quad \times \left( \frac{\Gamma(\beta)}{\Gamma(\beta + N)} \prod_{j=1}^{J} \frac{\Gamma(\beta_j + c_{+j})}{\Gamma(\beta_j)} \right) \\ &\equiv \Upsilon(\mathbf{c}_{I+}, \alpha_I) \Upsilon(\mathbf{c}_{+J}, \beta_J) \quad (6) \end{aligned}$$

again with $\alpha = \sum \alpha_i$ and $\beta = \sum \beta_j$. Given Eq. (4), (5) and (6), we arrive at our final formula of the posterior probability of independence at resolution $R$:

$$Pr(M_\mathcal{I} \mid \mathbf{D}) = 1 \bigg/ \left[ 1 + \frac{(1-\wp) \Upsilon(\mathbf{c}_K, \gamma_K)}{\wp \Upsilon(\mathbf{c}_{I+}, \alpha_I) \Upsilon(\mathbf{c}_{+J}, \beta_J)} \right]. \quad (7)$$

Eq. (7) refers to a fixed grid at resolution $I \times J$, not necessarily regular, with a given boundary set $\mathbf{B}_{I \times J}$.

In the next section, we will examine our data set at multiple resolutions. Intuitively speaking, we do not wish to use very fine ones due to data sparsity. This issue is automatically addressed by the above approach: for the choice of $\alpha_i = \beta_j = \gamma_k = 1$ (corresponding to a uniform Dirichlet prior), as $I$ and $J$ go to infinity, $Pr(M_\mathcal{I} \mid \mathbf{D}, \mathbf{B}_{I \times J}, I \times J)$ goes to unity. This can be seen as the effect of the hyperparameters, which represent "virtual samples." As the resolution increases the table becomes increasingly sparse and the actual counts tend to either 0 or 1. The effect of virtual samples then becomes apparent, making the table appear uniform by overwhelming the effect of the actual data. This is obviously an artificial phenomenon caused by our choice of prior, but it is useful in giving us an indication of when the resolution is disproportionately large to support our estimation. It also automatically adjusts for overly complex discretizations: in the fol-lowing section this effect is used to stop the coarse-to-fine discretization that is part of our proposed algorithm of estimating the probability of independence at multiple resolutions.

We note that the above test applies to categorical variables only. Random rearrangement of the rows and/or columns of the table will not affect the joint or the marginal counts and thus not affect $Pr(M_\mathcal{I} \mid \mathbf{D})$. However the resulting table counts appear more random, since the cell positions have been randomly changed, including the points that are contained in each of them. A more powerful test that takes into account the ordering and relative position of the points is the one described in the next section. This involves "sweeping" the discretization boundaries across the $XY$ plane (and the subspace of the variables of the conditioning set), and computing a Bayesian average of the results.

Lastly, we briefly address conditional independence: a conditional test $(X \perp Y \mid \mathbf{C})$ at a fixed resolution $I \times J \times C_1 \times C_2 \times \cdots \times C_{|\mathbf{C}|}$ is simply the product of probabilities of independence of $X$ and $Y$ for each $I \times J$ "slice" from $(1, 1, \ldots, 1)$ to $(C_1, C_2, \ldots, C_{|\mathbf{C}|})$. This is necessary because a conditional independence statement is equivalent to independence for all values of the conditioning set.

In the next section we describe a test that takes into account multiple resolutions.

### 2.2 Multi-resolution test

As we mentioned in the introduction, to estimate the posterior probability of independence we need to examine our data set at multiple resolutions. In this section we employ a Bayesian approach and average over the possible choices, weighed by their posterior:

$$Pr(M_\mathcal{I} \mid \mathbf{D}) = \sum_R Pr(R \mid \mathbf{D}) \int Pr(M_\mathcal{I} \mid \mathbf{B}_R, R, \mathbf{D}) Pr(\mathbf{B}_R \mid R, \mathbf{D}) d\mathbf{B}_R$$

The sum goes over all resolutions $I \times J$ (for the unconditional test case). For each such resolution $R$, the integral runs over all possible sets of discretization grid boundaries $\mathbf{B}_R$. The case of conditional test is handled analogously. The Bayesian approach, besides making the smallest number of unwarranted decisions, also possesses certain other advantages in our case; most notably it minimizes the possibility of spurious (in)dependencies that may occur due to an unfortunate choice of boundary placement.

To compute the inner integral we should ideally average over all possible histogram boundary placements along the $X$ and $Y$ axes. Lacking any other information, we assume a uniform prior distribution $Pr(\mathbf{B}_R \mid R)$ over grid boundary placement. Although theoretically the Bayesian integral runs over an infinity of possible such placements, given our data set we need only compute a finite number of them since many produce the same 2D histogram for our data. More specifically, we need only attempt boundary placements at each of the $(N-1)$ midpoints between successive data points along the $X$ and $Y$ axes, resulting in $(N-1)^2$ possible positions for each $XY$ boundary pair. An appropriate posterior $Pr(\mathbf{B}_R \mid R, \mathbf{D})$ for each such placement is one that uses the fraction of the 2D area between the two successive points along $X$ and $Y$ relative to the span of the entire set of data points in the $XY$ plane.

We now address the sum over all possible resolutions $R$. The number of such resolutions is countably infinite. However, the following crucial observation about the structure of the space of resolutions facilitates our computation:

> **Observation:** If $X$ and $Y$ are independent at resolution $2M \times 2M$, they are independent at resolution $M \times M$. In other words, $Pr[i_{XY}(M) \mid i_{XY}(2M)] = 1$ for all $M$, where $i_{XY}(M)$ denotes the statement that $X$ and $Y$ are independent at resolution $M \times M$.

It is straightforward to prove the validity of the above—it can be demonstrated by using a $2M \times 2M$ grid where each cell's probability is the product of the corresponding marginal cells, and by combining adjacent cells in $2 \times 2$ groups. A similar statement can be made for an $I \times J$ table. For simplicity, during the remainder of this section we shall assume square tables.

By definition, $X$ and $Y$ are independent if and only if they are independent at all resolutions. However, given the above observation, we have

$$\begin{aligned} Pr[i_{XY}(M) \wedge i_{XY}(2M)] &= \\ Pr[i_{XY}(M) \mid i_{XY}(2M)] Pr[i_{XY}(2M)] &= \\ Pr[i_{XY}(2M)] \end{aligned}$$

This implies that in order to ascertain independence, we do not need to sum over all resolutions; instead, we should examine our data set at a resolution as fine as possible. Unfortunately, because our data set size is finite, we can only obtain histograms of up to a finite resolution. Estimating the posterior probability of independence at a fine, limiting resolution is therefore an important part of our algorithm, which is presented below.

Our method begins with a coarse $2 \times 2$ grid and successively refines it by adding carefully selected boundaries along the axes. At each step, the set of boundaries

```
To compute Pr(M_I | D):

 1.   B_save ← ∅
 2.   p_max ← 0
 3.   t ← 1
 4.   do
 5.      B ← B_save
 6.      q_max ← 0
 7.      for x = { midpoints along X axis }
 8.         for y = { midpoints along Y axis }
 9.            B' ← B ∪ {(x, y)}
10.            w' ← Pr(B' | R)
11.            p(x, y) ← [1 - Pr(M_I | B', R, D)] × w'
12.            if p(x, y) > q_max then
13.               q_max ← p(x, y)
14.               B_save ← B'
15.      p(t) ← ∑_{x,y} p(x, y)
16.      if p(t) > p_max
17.         p_max ← p(t)
18.      t ← t + 1
19.   while the standard deviation of p(t) is large
20.   return 1 - p_max
```

Figure 2: Algorithm for computing the posterior probability of independence using an irregular multiresolution grid.

we admit—either an $XY$ pair or a set containing one boundary for each variable in the conditioning set—is the one that increases the probability of dependence the most. From each new, (possibly) irregular grid, we compute its posterior probability of dependence using Bayesian averaging over the possible single-boundary choices (using Eq. (7) for each resulting grid) and we either iterate or stop. Pseudocode of our procedure is shown in figure 2. Its typical use in situations where a binary dependent/independent decision is required would be to compare its output $Pr(M_I \mid D)$ to the prior probability of independence $\wp$ and decide "independent" if and only if $Pr(M_I \mid D) > \wp$.

To see why our algorithm attempts to maximize the posterior probability of dependence we need to examine another implication of the above observation:

$$\begin{aligned}
Pr[i_{XY}(M)] &= \\
Pr[i_{XY}(M) \wedge i_{XY}(2M)] &+ \\
Pr[i_{XY}(M) \wedge \neg i_{XY}(2M)] &= \\
Pr[i_{XY}(M) \mid i_{XY}(2M)] Pr[i_{XY}(2M)] &+ \\
Pr[i_{XY}(M) \mid \neg i_{XY}(2M)] Pr[\neg i_{XY}(2M)] &= \\
Pr[i_{XY}(2M)] &+ \\
Pr[i_{XY}(M) \mid \neg i_{XY}(2M)] Pr[\neg i_{XY}(2M)] &\geq \\
Pr[i_{XY}(2M)]
\end{aligned}$$

This implies that the probability of independence decreases or remains the same as we examine our data set at increasingly fine resolutions, or, equivalently, that the dependence probability is non-decreasing. Equality occurs in the case of independence at all resolutions. The above statement is also intuitively true: imagine for example the two limiting cases: a single cell, and the case where resolution tends to infinity at an ideal scenario where we have an infinite data set at our disposal. In the first case, the posterior independence probability is 1 irrespective of the data set. In the second, unless the data set is truly independent at every resolution, any discrepancies of the posterior probability of each cell from the product of the corresponding marginals will eventually emerge as we refine our discretization, causing the posterior independence probability to decline.

Combining the above observation with the previous one concerning the need to examine our data at as fine resolution as possible, we see that we need to maximize the posterior of dependence in an effort to discern dependence, as we progressively increase the effective resolution during a run of our algorithm. One concern however is that our data may appear increasingly artificially dependent at finer resolutions, a statement that might be unwarranted by our (sparse) data. The solution to this comes from our choice of a Dirichlet parameter prior ($\alpha_i = \beta_j = \gamma_k = 1$), which returns a high value of posterior independence at very high resolutions where there are very few data points and a large number of empty cells (see discussion in previous section). This corresponds well with our intuition, and lends itself to the estimation of the finest resolution to use: it is the one that returns a minimum for the posterior probability of independence. That posterior, under our choice Dirichlet prior, is "U-shaped," starting at 1 at resolution $1 \times 1$, decreasing and then tending to 1 again at high resolutions.

The above statement about the non-increasing probability of independence as we add boundaries refers to $XY$ boundaries only. Unfortunately it is not the case for adding boundaries along the conditioning set variable axes. It is not difficult to construct artificial examples where that may be demonstrated. Due to lack of space, we do not present such examples here. But the net effect is that increasing the number of conditioning values can increase or decrease the probability of independence, as judged by the resulting multidimensional histogram. Investigating the behavior with respect to conditioning is one of the future research directions we intend to follow.

An example run of our algorithm on two data sets, one independent by construction and one where $X$ and $Y$ exhibit a highly nonlinear dependency is shown in

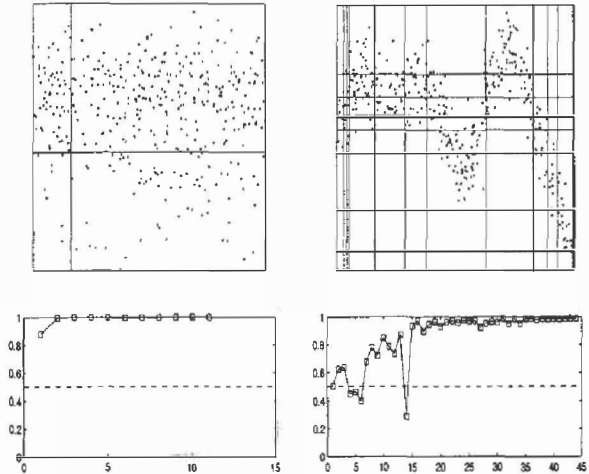

Figure 3: An example run of our algorithm on two data sets, the same ones as in figure 4. **Left:** Data independent. **Right:** Data nonlinearly dependent. **Top scatterplots:** The actual data set, together with the grid that is output from our algorithm, drawn at the point of minimum independence. **Bottom plots:** The probability of independence as boundaries are added to the grid. The dashed horizontal line represents our prior $\wp = 0.5$.

figure 3. In figure 3, bottom, we plot the evolution of $p(t)$ as the $XY$ is discretized at successively finer resolutions. Both plots demonstrate how the posterior probability of independence tends to unity as the grid becomes more complex. The plot on the right is not strictly "U-shaped" due to the fact that subdivision steps do not exactly double the resolution. The top plots show the grid that produces the discretization of maximum dependence, i.e. $Pr(M_{\mathcal{I}} \mid \mathbf{D}) = 1 - p_{max}$. Note that the set of independent points contains only four cells. Our algorithm returns posterior probability of independence $1 - p_{max} = 0.88$ for the independent data set and 0.29 for the dependent one.

The algorithm of figure 2 essentially computes an approximation of the complete Bayesian integral over all possible grid boundaries at the resolution corresponding to the minimum probability of independence supported by our data set. At this limiting resolution $M^* \times M^*$, we treat the grid boundaries in a Bayesian fashion: ideally we should go over all possible $M^* \times M^*$ grids whose boundaries lie on each possible pair of midpoints between successive data points along the axes. However, the number of such irregular $M^* \times M^*$ grids is very large, namely $\binom{N-1}{M^*-1}^2$, a number exponential in the number of data points. Our approach approximates this computation by maximizing over the first $M^* - 1$ boundaries and averaging over the $M^*$-th one. This results in a polynomial-time approximation of the Bayesian sum.

In summary, the justification of our algorithm is as follows: as we established by the structure of the resolution space, one needs to minimize posterior independence across resolutions, while approximating the average over an exponential sum over possible boundaries at each such resolution. Our algorithm accomplishes both goals simultaneously by maximizing dependence instead of minimizing independence across resolutions by employing an incremental maximum-value approximation to the evaluation of successive Bayesian integrals over multiple resolutions. In doing so, it reduces an exponential computation to a polynomial-order one.

## 3 Experimental results

In this section we compare our approach with existing ones in use routinely by statisticians. Perhaps the most frequently used method for non-parametric independence testing is rank correlation, either Spearman's rank order correlation coefficient $r_s$ and Kendall's tau ($\tau$) [9]. The two measures are related, and result in essentially the same test in most cases. Here we will compare our test with Spearman's coefficient as it is the most expressive of the two. We will use some examples from the Boston housing data set, which is contained in the UC Irvine data set repository [7].

In figure 4, we see two variables that appear dependent, the average number of rooms per house and its median value. The dependence between them is deemed very weak by rank correlation (its correlation coefficient is only 0.08), while our algorithm returns 0.99 as the posterior probability of dependence. We see the evolution of that probability for the first few steps in the bottom plot of figure 4. These variables were found dependent in the original study.

Some other examples of dependent (figure 5, rank correlation is -0.02 while our algorithm returns 0.98) and approximately independent pairs of variables (figure 6, rank correlation 0.01, posterior dependence 0.32, both indicating independence) demonstrate how our approach works as expected in practice, and better than existing established techniques.

## 4 Related Work

The most well-known independence test is perhaps the chi-square ($\chi^2$) test. It operates on categorical data, and assumes there are enough counts in each bin such that the counts are approximately normally distributed. Addressing the independence in $2 \times 2$ con-tingency tables, Fisher proposed an exact test in 1934 (see [1] for one description). The difference between ours and Fisher's exact test is that the latter one is not Bayesian, and it computes a power value rather than the posterior probability. Bayesian approaches to contingency table independence also exist: the $2 \times 2$ case is addressed in Jeffreys' classic text [4], and in general more recently (e.g. [3]), although not in exactly the same way as ours. However the greatest difference that makes our method unique is that it applies to continuous variables, taking advantage of the ordering of the values.

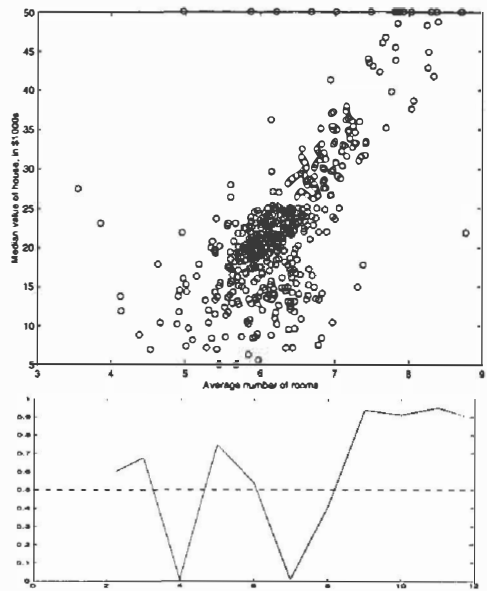

Figure 4: Two dependent variables from the Boston housing data set, the average number of rooms per dwelling and and the median value of owner-occupied homes, in thousands of dollars. Spearman's rank correlation is very small (0.081) due to the fact that it is a two-dimensional distribution. In the bottom we plot the posterior independence probability during a run of our algorithm. The resulting posterior dependence probability is 0.99.

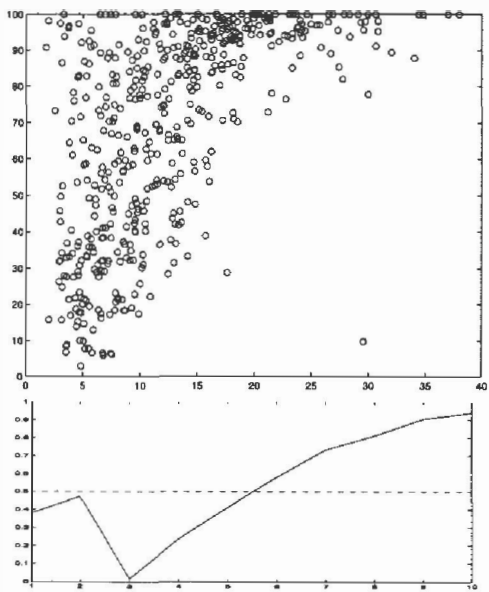

Figure 5: Percent lower status of population vs. proportion of houses built before 1940. Rank correlation indicates independence with a value of -0.02, while our method suggests strong dependence (0.98).

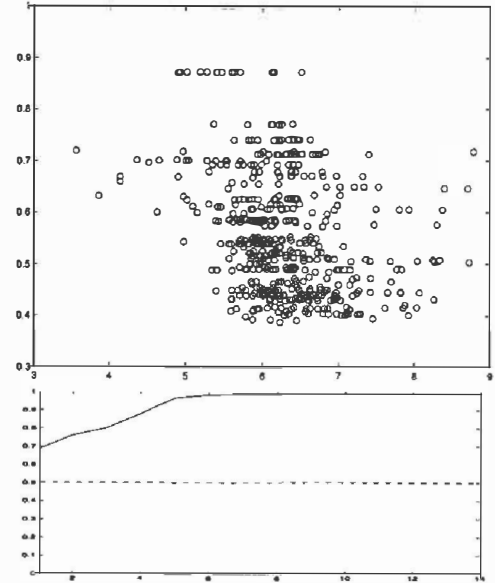

Figure 6: On the $X$ axis, the average number of rooms. On the $Y$ axis, an air-pollution indicator. The variables are deemed independent both by Spearman's rank correlation coefficient (0.01) and our method (0.32).

The most well-known statistical non-parametric tests of independence for ordinal variables are Spearman's rank correlation and Kendall's tau. The former is the linear correlation of the relative ranks of the points along the $X$ and $Y$ axes, while the latter reduces ranks to binary or ternary variables (larger, equal, or smaller). Both tests can handle clear monotonic trends but fail to capture complex interactions between $X$ and $Y$. Most notably, as demonstrated in section 3, they also fail to capture joint probability functions of dependent variables that are two dimensional.

As far as parametric approaches are concerned, there are a multitude of such approaches as they are the norm in practice. There are two main issues with all these approaches: first, they assume a certain family of distributions, which may fail in cases where the underlying data distribution does not fit well. Second, the question of proper bias and its tradeoff with variance (the issue of model complexity) needs to be adequately addressed. This problem also occurs in our approach, and is solved though our choice of parameter prior.

## 5 Conclusion and Future Research

In this paper we presented a probabilistic test of conditional independence between two continuous variables. For this we proposed the posterior probability of independence given the data, taking into account histograms representing discretizations at several different resolutions. We show that this is necessary by observing that the range of values that this quantity may take across different resolutions may vary substantially. We developed a Bayesian approach, integrating over all possible boundary placements on the plane, and incorporating evidence from different resolutions. By taking advantage of certain structure that exists in the space of discretizations, we were able to approximate this quantity, reducing the time that is required to estimate it from exponential in the size of the data set, to polynomial.

An issue that merits further attention is the case of the conditional test. Our proposed extension to the unconditional test may not work in all cases, because the observation of section 2.2 does not extent to the conditioning set variables, as is explained later in that section (it may work as an approximation). Therefore a more general algorithm is needed for arbitrarily distributed data. We plan to work on this issue in the near future.

Our solution may benefit the generation of Bayesian networks in domains that contain any number of continuous, ordinal discrete and/or categorical attributes. In such domains, our test may be applied in a straightforward fashion, due to the fact that we employ discretization. One additional benefit that may emerge from such an approach is the representation of the conditional probability distributions of a BN, if our test is being used for such a purpose. This is another topic of further research.

Another interesting direction of research is the development of quad-tree like discretization rather than axis-parallel boundaries that span the range of the variables involved. One can envision, for example, a discretization that detects ranges in the plane where the variables are locally approximately independent. Such a range may indicate the presence of another quantity, explaining the local independence. This may be used to generate new, possibly meaningful features by combining existing ones in the domain.

Finally, accelerating the asymptotic running time of our algorithm is important. Even though our method is polynomial in time, the exponent is the number of variables involved in the conditional test. This may be prohibitively expensive for large conditioning sets and/or data sets. A faster algorithm is required for those situations.

## Acknowledgements

We wish to thank Greg Cooper for his helpful comments during the development of the method and theory behind this paper and some useful pointers to relevant literature.